\documentclass{article}
\IfFileExists{KLMM/klmm.sty}{%
  \usepackage[]{KLMM/klmm}%
}{%
  \usepackage[margin=1in]{geometry}%
  \usepackage{caption}%
}
\usepackage{booktabs}
\usepackage{tikz}
\usepackage{pgfplots}
\usepackage[T1]{fontenc}
\usepackage{tabularx}
\usepackage{xurl}
\usepackage{hyperref} % Omit if already loaded by the template.
\usepackage{lmodern}
\usepackage[scaled=0.95]{helvet}
\usepackage{graphicx}
\pgfplotsset{compat=1.18}
\usepackage[hidelinks]{hyperref}
\usepackage{tcolorbox}
\usepackage{subcaption}
\usepackage{amsmath, amssymb}
\tcbuselibrary{skins,breakable}
\usepackage{listings}
\usepackage{float}  % 必须放在下面这些命令之前

\floatstyle{ruled}
\newfloat{listing}{tb}{lst}
\floatname{listing}{Listing}
\floatstyle{plain}
\newfloat{reviewcase}{tbp}{lor}
\floatname{reviewcase}{Case}
\DeclareCaptionStyle{ruled}{labelfont=normalfont,labelsep=colon,strut=off} % DO NOT CHANGE THIS
\usepackage{amsfonts}
\usepackage{amsopn}
\definecolor{keywordcolor}{rgb}{0.7, 0.1, 0.1}
\definecolor{tacticcolor}{rgb}{0.0, 0.1, 0.6}
\definecolor{commentcolor}{rgb}{0.4, 0.4, 0.4}
\definecolor{symbolcolor}{rgb}{0.0, 0.1, 0.6}
\definecolor{sortcolor}{rgb}{0.1, 0.5, 0.1}
\definecolor{stringcolor}{rgb}{0.5, 0.3, 0.0}

\lstnewenvironment{leancode}[1][]{%
  \lstset{language=lean}%
  \lstset{mathescape=false,extendedchars=true}%
  #1%
}{}

\DeclareUnicodeCharacter{22A4}{\ensuremath{\top}}
\DeclareUnicodeCharacter{2227}{\ensuremath{\wedge}}

\definecolor{informalaccent}{HTML}{C96F68}
\definecolor{formalaccent}{HTML}{6F9BC6}

\tcbset{claimbox/.style={
    enhanced,
    colback=white,
    boxrule=0.5pt,
    sharp corners,
    boxsep=3pt,
    left=3pt,
    right=3pt,
    top=10pt,
    bottom=3pt,
    before skip=0pt,
    after skip=0pt,
    overlay={
        \node[anchor=north east,inner sep=0pt,font=\scriptsize,
              text=black!60]
        at ([xshift=-6pt,yshift=-3pt]frame.north east) {#1};
    }
}}

\newtcolorbox{informalbox}{claimbox=Informal,colframe=black}
\newtcolorbox{formalbox}{claimbox=Formal,colframe=black,
    before skip=-0.5pt,
    before upper={\lstset{aboveskip=0pt,belowskip=0pt}}}

\title{MechAlign: Evaluating Semantic Alignment at the Project Level in Mathematical Formalization}
\author{Hao~Shen\institution{State Key Laboratory of Mathematical Sciences, Academy of Mathematics and Systems Science, Chinese Academy of Sciences, and University of Chinese Academy of Sciences. }\equalcontrib
    \and Ruichen~Qiu\instref{1}\equalcontrib
    \and Lihong~Zhi\instref{1}
    \and Xiao-Shan~Gao\instref{1}
}
\newcommand{\CAT}{\mathrm{CAT}}

\newcommand{\bench}{LeanReviewBench}

\begin{document}
\maketitle

\begin{abstract}
Lean verifies that a formal proof establishes its stated proposition under declared assumptions, but does not check whether that proposition faithfully expresses the mathematical claim in a manuscript.
Assessing this correspondence requires interpreting the main theorem through its project dependencies, which may be scattered throughout a large development.
We introduce MechAlign, a framework for project-level semantic alignment review that combines formal dependency extraction with LLM-based assessment.
Its extraction tool, LeanTrace, constructs a statement dependency graph from elaborated declarations, retaining definition bodies and structure constraints while excluding proof bodies and stopping at an explicit library boundary.
The reviewer uses this context to investigate alignment and report discrepancies with localized evidence.
We conduct a preliminary review of 201 main results from 173 OpenAI Math projects, using supplied informal descriptions and available paper context.
A hyperbolic-group case illustrates how dependency inspection localizes a missing dimensional constraint relative to the manuscript, consistent with the upstream declaration of formalization scope.
We also construct LeanReviewBench, comprising 501 fixed claim–statement–context triples from OpenAI Math, Fermat's Last Theorem, and Kakeya, with description provenance recorded.
The resource supports controlled comparisons of more project-level semantic alignment evaluations.
%statement-only, full-project, and dependency-guided review; comparative results and human adjudication remain pending.
\end{abstract}

\section{Introduction}
\label{sec:introduction}
Recent advances in large language models have enabled the generation of mathematical proofs at increasing scale~\citep{cao2026mechmath, liu2026danus, zhang2026continual,openai2026math}.
Lean formalization is also used to verify the stated results. For example, OpenAI's mathematical research collection includes Lean projects for 173 manuscripts~\citep{openai2026math}.
% These proofs are increasingly accompanied by Lean formalizations intended to verify their claims.
% For example, OpenAI's mathematical research collection, includes accompanying Lean projects for 173 manuscripts~\citep{openai2026math}.
Lean checks that a formal proof establishes its stated proposition under the declared assumptions~\citep{moura2021lean}, but it does not check whether that proposition faithfully expresses the mathematical claim in the manuscript.
This leaves a semantic gap:
\begin{quote}
    \em Does the statement proved in the Lean project faithfully express the mathematical claim made in the manuscript?
\end{quote}
Answering this question requires establishing the mathematical meaning of the formal statement.
In a Lean project, that meaning is determined not only by the main theorem declaration but also by the definitions and constraints it refers to.

% For example, a theorem stated using a project-defined predicate may appear to match the manuscript, while the predicate's definition imposes additional assumptions or specifies a weaker conclusion.
% Interpreting these declarations may in turn require examining their dependencies.
% We therefore use \emph{project-level alignment review} to mean assessing a selected formal result against its informal claim through the project declarations that determine its meaning.
For example, a theorem stated using a predicate defined within the project may appear to match the manuscript, even though the predicate encodes additional assumptions or a weaker conclusion.
Interpreting the predicate may, in turn, require tracing its dependencies.
We use \emph{project-level alignment review} to mean assessing a selected formal result against its informal claim by examining the project declarations that determine the result's meaning.

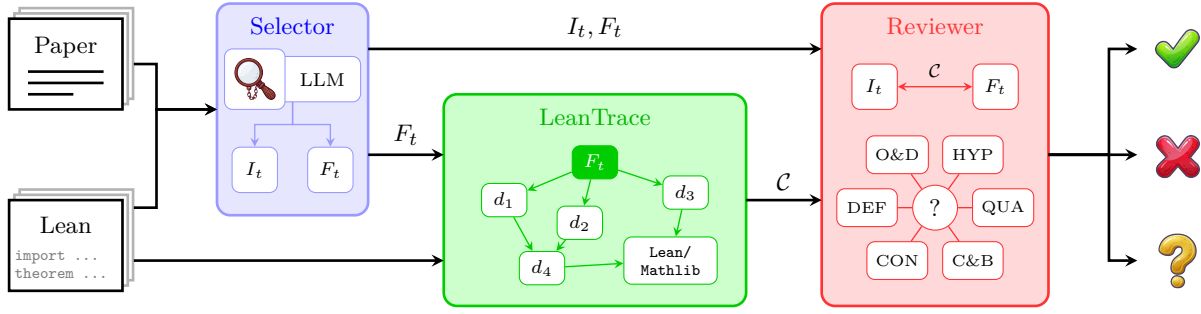
\begin{figure}[t]
\centering
\begin{tikzpicture}[
    font=\small, align=center
]
    \node at (-6.86,1.34) [rectangle, draw=black!30, fill=white, minimum width=1.5cm, minimum height=1.2cm,line width=1pt] {};
    \node at (-6.93,1.27) [rectangle, draw=black!30, fill=white, minimum width=1.5cm, minimum height=1.2cm,line width=1pt] {};
    \node at (-7,1.2) [rectangle, draw=black, fill=white, minimum width=1.5cm, minimum height=1.2cm,line width=1pt] {};
    \node at (-7,1.43) {Paper};
    \draw [-,draw=black,line width=1pt] (-7.5,1.1) -- (-6.5,1.1);
    \draw [-,draw=black,line width=1pt] (-7.5,0.95) -- (-6.5,0.95);
    \draw [-,draw=black,line width=1pt] (-7.5,0.8) -- (-6.9,0.8);

    \node at (-6.86,-1.06) [rectangle, draw=black!30, fill=white, minimum width=1.5cm, minimum height=1.2cm,line width=1pt] {};
    \node at (-6.93,-1.13) [rectangle, draw=black!30, fill=white, minimum width=1.5cm, minimum height=1.2cm,line width=1pt] {};
    \node at (-7,-1.2) [rectangle, draw=black, fill=white, minimum width=1.5cm, minimum height=1.2cm,line width=1pt] {};
    \node at (-7,-0.92) {Lean};
    \node at (-7,-1.45) [text width=1.35cm, align=left] {\textcolor{black!50}{\tiny\ttfamily import ...\\theorem ...\\}};

    \node at (-4,0.6) [rectangle, rounded corners=5pt, draw=blue!40, fill=blue!10, minimum width=2cm, minimum height=2.8cm, line width=1pt] {};
    \node at (-4,1.7) {\textcolor{blue}{Selector}};
    \node at (-3.7,1) [rectangle, rounded corners=3pt, draw=blue!40, fill=white, minimum width=1.2cm, minimum height=.6cm] {\scriptsize ~~LLM};
    \node at (-4.5,1) [rectangle, rounded corners=3pt, draw=blue!40, fill=white, minimum width=.8cm, minimum height=.8cm] {};
    \node at (-4.5,1) {\includegraphics[width=.8cm]{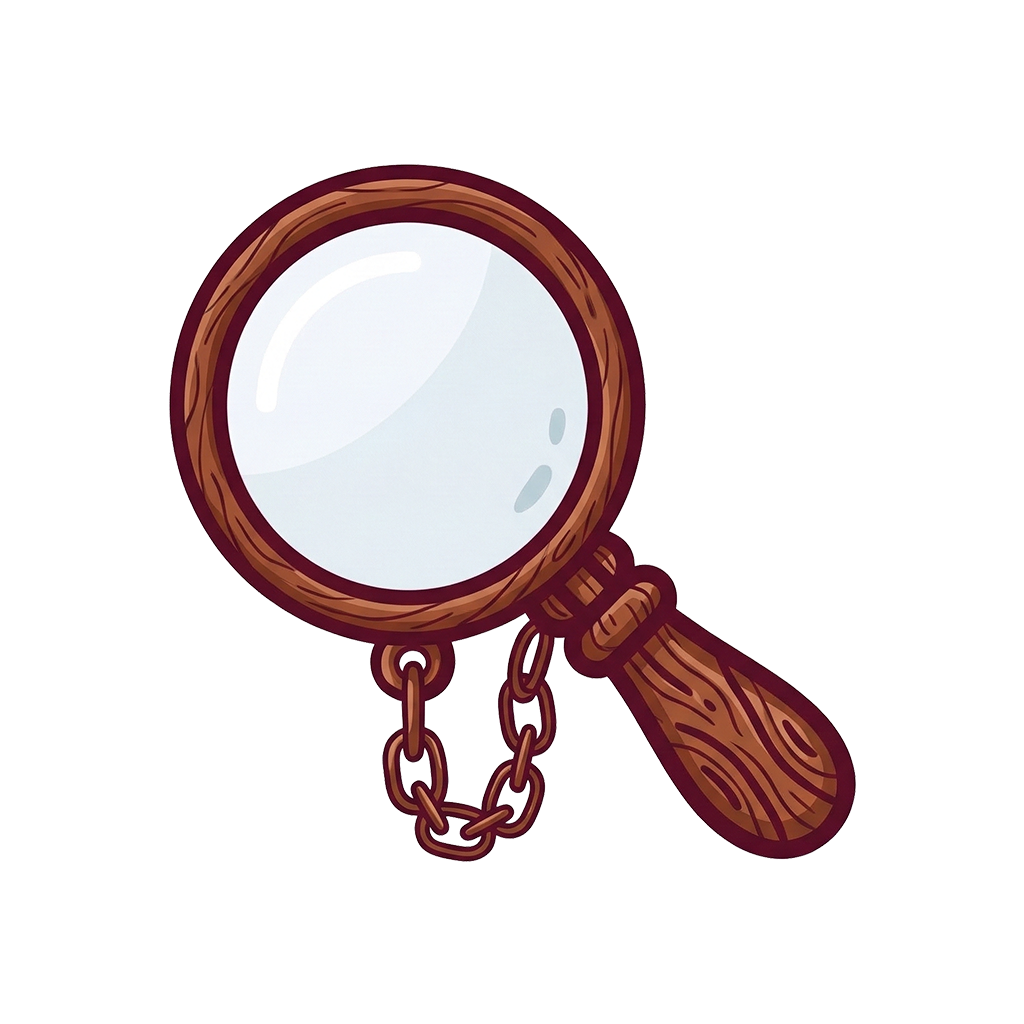}};
    \node (i_0) at (-4.5,-0.2) [rectangle, rounded corners=3pt, draw=blue!40, fill=white, minimum width=.6cm, minimum height=.6cm] {\scriptsize $I_t$};
    \node (f_0) at (-3.5,-0.2) [rectangle, rounded corners=3pt, draw=blue!40, fill=white, minimum width=.6cm, minimum height=.6cm] {\scriptsize $F_t$};
    \draw [->, >=stealth, line width=0.5pt, draw=blue!40] (-4,0.7) --++ (0,-0.3) -| (i_0.north);
    \draw [->, >=stealth, line width=0.5pt, draw=blue!40] (-4,0.7) --++ (0,-0.3) -| (f_0.north);

    \node at (0,-0.6) [rectangle, rounded corners=5pt, draw=green!80!black, fill=green!20, minimum width=4cm, minimum height=2.8cm, line width=1pt] {};
    \node (d_0) at (0,-0.1) [rectangle, rounded corners=3pt, draw=green!80!black, fill=green!80!black, minimum width=.6cm, minimum height=.4cm] {\scriptsize\textcolor{white}{$F_t$}};
    \node (d_1) at (-1.2,-.6) [rectangle, rounded corners=3pt, draw=green!80!black, fill=white, minimum width=.6cm, minimum height=.4cm] {\scriptsize $d_1$};
    \node (d_2) at (-0.2,-0.9) [rectangle, rounded corners=3pt, draw=green!80!black, fill=white, minimum width=.6cm, minimum height=.4cm] {\scriptsize $d_2$};
    \node (d_3) at (1.2,-.5) [rectangle, rounded corners=3pt, draw=green!80!black, fill=white, minimum width=.6cm, minimum height=.4cm] {\scriptsize $d_3$};
    \node (d_4) at (-.7,-1.5) [rectangle, rounded corners=3pt, draw=green!80!black, fill=white, minimum width=.6cm, minimum height=.4cm] {\scriptsize $d_4$};
    \node (d_5) at (1,-1.4) [rectangle, rounded corners=3pt, draw=green!80!black, fill=white, text width=1cm] {\tiny\ttfamily Lean/\\Mathlib\\};
    \draw [->, >=stealth, line width=0.5pt, draw=green!80!black] (d_0) -- (d_1);
    \draw [->, >=stealth, line width=0.5pt, draw=green!80!black] (d_0) -- (d_2);
    \draw [->, >=stealth, line width=0.5pt, draw=green!80!black] (d_0) -- (d_3);
    \draw [->, >=stealth, line width=0.5pt, draw=green!80!black] (d_1) -- (d_4);
    \draw [->, >=stealth, line width=0.5pt, draw=green!80!black] (d_2) -- (d_4);
    \draw [->, >=stealth, line width=0.5pt, draw=green!80!black] (d_3) -- (d_5);
    \draw [->, >=stealth, line width=0.5pt, draw=green!80!black] (d_4) -- (d_5);
    \node at (0,0.5) {\textcolor{green!70!black}{LeanTrace}};

    \node at (4.5,0) [rectangle, rounded corners=5pt, draw=red!80, fill=red!15, minimum width=3cm, minimum height=4cm, line width=1pt] {};
    \node at (4.5,1.7) {\textcolor{red!80}{Reviewer}};
    \node (i_1) at (3.7,.9) [rectangle, rounded corners=3pt, draw=red!80, fill=white, minimum width=.6cm, minimum height=.6cm] {\scriptsize $I_t$};
    \node (f_1) at (5.3,.9) [rectangle, rounded corners=3pt, draw=red!80, fill=white, minimum width=.6cm, minimum height=.6cm] {\scriptsize $F_t$};
    \draw [<->, >=stealth, line width=0.5pt, draw=red!80] (i_1) -- node[above]{\scriptsize $\mathcal{C}$} (f_1);
    
    \node (od) at (4,0) [rectangle, rounded corners=3pt, draw=red!80, fill=white, minimum width=.8cm, minimum height=.5cm] {\tiny O\&D};
    \node (hyp) at (5,0) [rectangle, rounded corners=3pt, draw=red!80, fill=white, minimum width=.8cm, minimum height=.5cm] {\tiny HYP};
    \node (qua) at (5.4,-0.7) [rectangle, rounded corners=3pt, draw=red!80, fill=white, minimum width=.8cm, minimum height=.5cm] {\tiny QUA};
    \node (cb) at (5,-1.4) [rectangle, rounded corners=3pt, draw=red!80, fill=white, minimum width=.8cm, minimum height=.5cm] {\tiny C\&B};
    \node (con) at (4,-1.4) [rectangle, rounded corners=3pt, draw=red!80, fill=white, minimum width=.8cm, minimum height=.5cm] {\tiny CON};
    \node (def) at (3.6,-0.7) [rectangle, rounded corners=3pt, draw=red!80, fill=white, minimum width=.8cm, minimum height=.5cm] {\tiny DEF};
    \node (question) at (4.5,-0.7) [circle, draw=red!80, fill=white, minimum width=.5cm] {?};
    \draw [-, line width=0.5pt, draw=red!80] (od) -- (question);
    \draw [-, line width=0.5pt, draw=red!80] (hyp) -- (question);
    \draw [-, line width=0.5pt, draw=red!80] (qua) -- (question);
    \draw [-, line width=0.5pt, draw=red!80] (cb) -- (question);
    \draw [-, line width=0.5pt, draw=red!80] (con) -- (question);
    \draw [-, line width=0.5pt, draw=red!80] (def) -- (question);

    \node (al) at (7.7,1.4) {\includegraphics[width=.8cm]{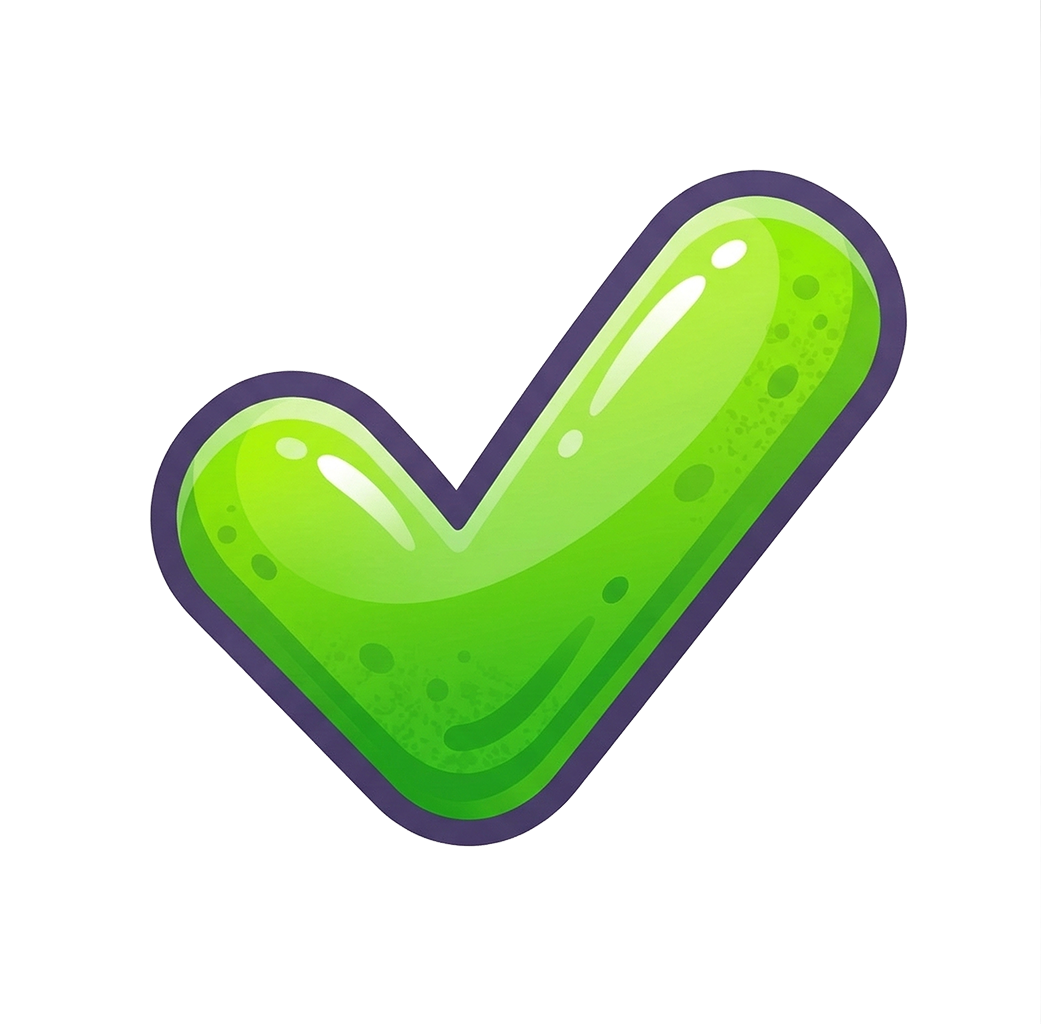}};
    \node (misal) at (7.7,0) {\includegraphics[width=.8cm]{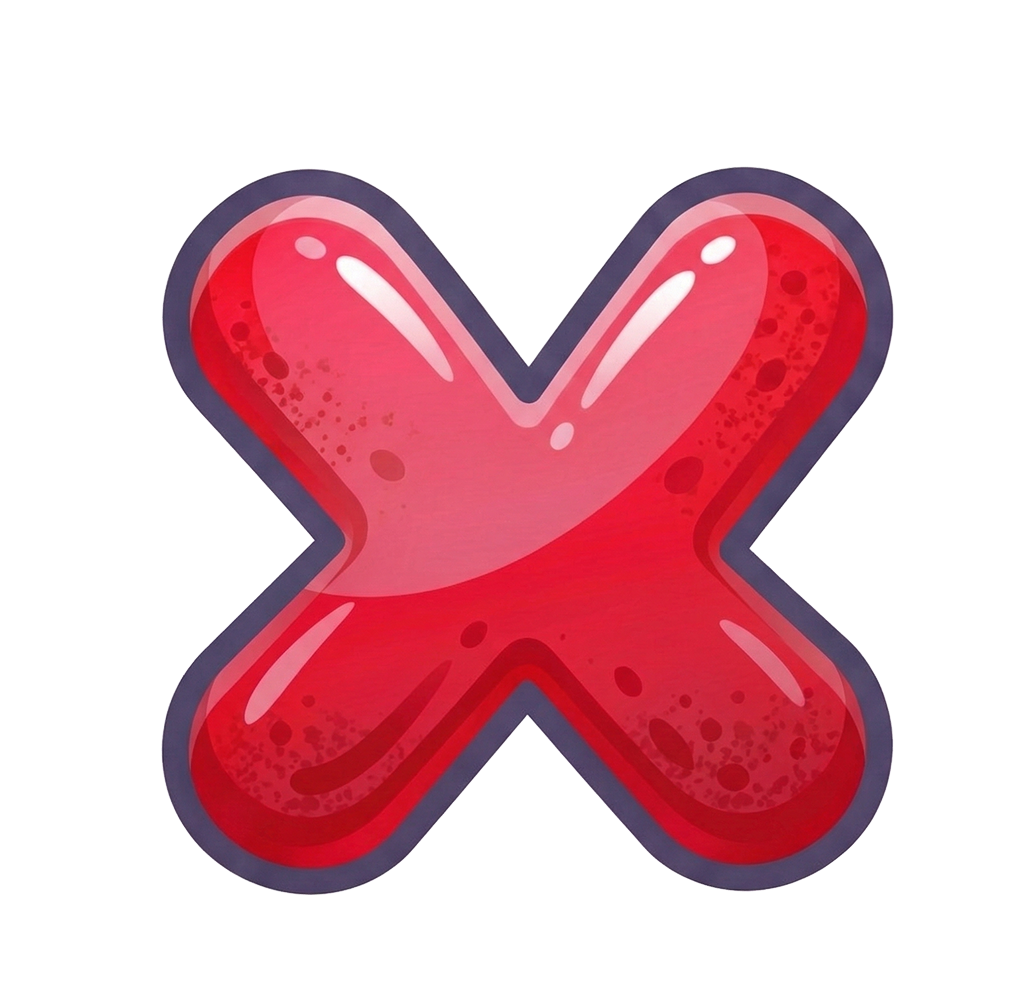}};
    \node (und) at (7.7,-1.4) {\includegraphics[width=.8cm]{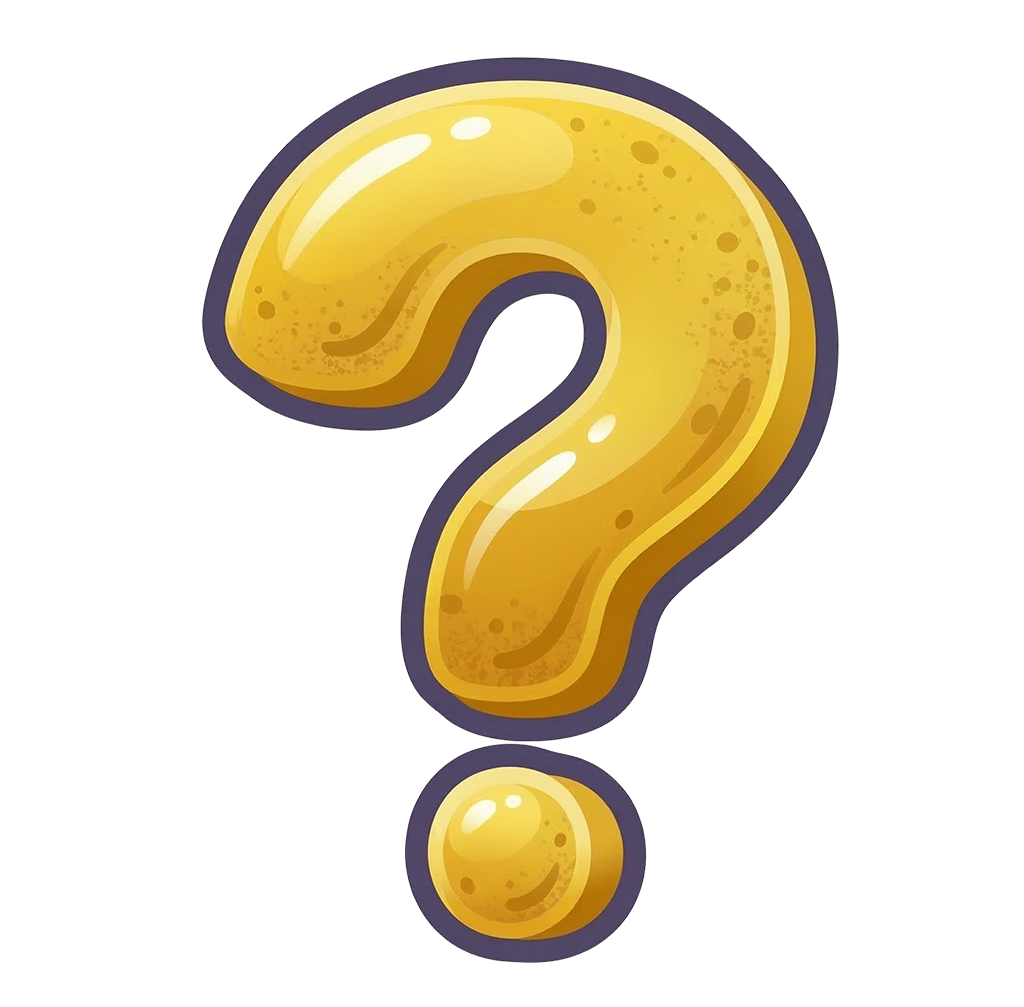}};
    
    \draw [->, >=stealth, line width=1pt, draw=black] (-6.1,1.2) --++ (0.3,0) |- (-5,0.6);
    \draw [->, >=stealth, line width=1pt, draw=black] (-6.1,-0.7) --++ (0.3,0) |- (-5,0.6);
    \draw [->, >=stealth, line width=1pt, draw=black] (-3,0) -- node[above]{$F_t$} (-2,0);
    \draw [->, >=stealth, line width=1pt, draw=black] (-6.1,-1.4) -- (-2,-1.4);
    \draw [->, >=stealth, line width=1pt, draw=black] (-3,1.4) -- node[above]{$I_t,F_t$} (3,1.4);
    \draw [->, >=stealth, line width=1pt, draw=black] (2,-0.6) -- node[above]{$\mathcal{C}$} (3,-0.6);
    \draw [->, >=stealth, line width=1pt, draw=black] (6,0) --++ (0.8,0) |- (al.west);
    \draw [->, >=stealth, line width=1pt, draw=black] (6,0) --++ (0.8,0) |- (misal.west);
    \draw [->, >=stealth, line width=1pt, draw=black] (6,0) --++ (0.8,0) |- (und.west);
    
\end{tikzpicture}
\caption{Overview of MechAlign. The Selector obtains the paper and formal statements, LeanTrace extracts the formal dependency context, and the Reviewer assesses alignment using both formal and paper context.}
\label{fig:framework}
\end{figure}

Several methods can be used for project-level alignment review.
A \emph{statement-only} reviewer sees the main Lean declaration, but cannot inspect these supporting declarations.
Providing \emph{full project access} makes them available, but leaves the reviewer to locate and connect the relevant evidence.
We therefore ask how to organize these declarations to support both evidence gathering and mathematical assessment.
We introduce \textbf{MechAlign}, a framework for project-level semantic alignment review. Its formal evidence extractor, \textbf{LeanTrace}, starts from a selected theorem's elaborated type and follows references through declaration types and non-proof definition bodies.
% It retains structural constraints and stops expansion at an explicit library boundary, producing a statement dependency graph.
% MechAlign's LLM-based Reviewer inspects declarations and follows graph edges to assess the informal–formal correspondence and report discrepancies with their supporting evidence.
% This provides a third review strategy in which formal evidence is organized around the selected claim before semantic assessment.
% The extracted context supports investigation without guaranteeing that every mathematical interpretation can be resolved within it.
It retains structural constraints and stops expansion at an explicit library boundary, producing a statement dependency graph that organizes formal evidence around the selected result.
MechAlign's LLM-based reviewer inspects declarations and follows graph edges to assess alignment between the informal claim and the formal result, reporting discrepancies with supporting evidence.
The extracted context supports this assessment but may not suffice to resolve every question of mathematical interpretation.
Refer to Figure~\ref{fig:framework} for the structure of MechAlign. 
%Code and data release locations are provided in Appendix~\ref{app:code-availability}.

Our primary application is OpenAI Math~\citep{openai2026math}, where we inspect 201 selected main results from 173 projects and identify candidate discrepancies for further review.
The HyperbolicObstruction case illustrates how inspecting a statement and its supporting structure reveals that the manuscript's two-dimensionality requirement is not explicitly retained, consistent with the upstream declaration of formalization scope.
We extend the extracted contexts with 200 selected targets from Fermat's Last Theorem~\citep{anthropic2026formalizingflt} and 100 from Kakeya~\citep{numina2026kakeya} to construct \textbf{LeanReviewBench}, a resource of 501 fixed instances.
% Each instance pairs a supplied informal claim with a selected formal statement and its extracted dependency context, enabling reproducible comparisons of evidence access and semantic judgment.
% The supplementary collections broaden mathematical and dependency-structure coverage, but their Lean-derived descriptions support supplied-text consistency studies rather than independently establishing fidelity to a manuscript.
% We define a controlled comparison of statement-only, full-project, and dependency-guided review to assess discrepancy detection, evidence localization, and review cost; comparative results and human adjudication remain to be completed.
Each instance pairs a supplied informal claim with a selected formal statement and includes the statement's extracted dependency context. 
This structure supports a controlled, reproducible comparison of statement-only, full-project, and dependency-guided review in terms of discrepancy detection, evidence localization, and review cost. 
Supplementary collections broaden coverage of mathematical topics and dependency structures. 
Because their descriptions are derived from Lean, they support consistency checks against the supplied text but do not independently establish fidelity to a manuscript. 
The comparative evaluation and human adjudication remain to be completed.

Our contributions are threefold:
\begin{itemize}
    \item We formulate a project-level semantic alignment review in which a selected formal result is assessed against its informal claim through the declarations that determine its meaning.
    \item We introduce MechAlign, combining LeanTrace's extraction of statement dependencies with an LLM-based reviewer that inspects the resulting graph and reports localized semantic discrepancies.
    \item We construct LeanReviewBench, comprising 501 review instances from OpenAI Math, Fermat's Last Theorem, and Kakeya, and we define a controlled comparison of statement-only, full-project, and dependency-guided review.
\end{itemize}

\section{Related Work}
\paragraph{Autoformalization}
Autoformalization converts informal problems into formal statements that can be verified in Lean~4~\citep{moura2021lean}.
Herald~\citep{gao2025herald} annotates Mathlib~4 with natural language and trains a dedicated Lean~4 translator.
Process-driven Autoformalization~\citep{lu2024process} incorporates Lean compiler feedback through process-supervised verification.
SAIR-Comb~\citep{jiang2026sair} combines structure-aware refinement with expert iteration and reinforcement learning for autoformalization in combinatorics.
Agent-based approaches further integrate generation, checking, and refinement into interactive workflows.
MASA~\citep{zhang2025masa} coordinates agents for formalization, syntactic checking, semantic assessment, and iterative repair.
LeanMarathon~\citep{zhang2026leanmarathon} uses an evolving blueprint to coordinate agents across research-level formalization tasks.
Numina-Lean-Agent~\citep{liu2026numinaleanagent} combines a general-purpose coding agent with Lean interaction and theorem retrieval tools through MCP.
MechMath~\citep{qiu2026mechmath} uses sorrifier-driven decomposition to isolate unresolved subgoals while preserving the surrounding verified proof structure.
Numina Fuse~\citep{numina2026fuse} provides a desktop workspace for developing Lean formalizations with AI agents. These advances increase the need to review completed formalizations.
% Our focus is on checking an existing claimed result through its project dependencies, rather than generating or repairing its proof.

\paragraph{Evaluation Metrics for Faithfulness}
FormalAlign~\citep{lu2025formalalign} constructs evaluation sets from FormL4 and miniF2F~\citep{zheng2021minif2f} by augmenting existing informal-formal pairs with synthetic misalignments.
The Human Equivalence Benchmark~\citep{liu2025rethinking} provides 200 human-labeled candidate-reference pairs, while ProofNetVerif~\citep{poiroux2025reliable} contains 3,752 annotated examples for evaluating both reference-based and reference-free metrics.
MA-Align~\citep{patel2026mathatlas} extends evaluation to graduate-level mathematics through 200 type-correct formalizations of statements and definitions labeled for semantic alignment.

\paragraph{Semantic Alignment Evaluation}
Recent work further examines diagnostic detail and contextual dependencies.
FormalRx~\citep{wang2026formalrx} evaluates alignment judgments together with error categorization, localization, and correction. FaithAlign~\citep{feng2026aligning} introduces phrase-level correspondence annotations with explicit informal context and formal dependencies, while FaithAlign-Gap tests whether evaluators recognize insufficient context.
These efforts motivate studying evidence acquisition as part of alignment review.
LeanTrace supplies an explicit extraction policy for the statement dependencies of a selected result, and the review agent inspects that graph for semantic discrepancies.
Our full-project comparison is intended to test the value of this organization of evidence; the method does not presume that source retrieval or existing alignment systems necessarily fail.

\section{Method}
\label{sec:method}
In this section, we first define the project-level semantic alignment task. We then present LeanTrace, a tool for extracting the dependency graph of a selected theorem in a Lean project. Finally, we introduce MechAlign, an agent that uses this graph to evaluate semantic alignment in project-level mathematical formalization.

\subsection{Task Definition}
\label{sec:task}
For a selected review target in a Lean project, let $I_t$ denote the supplied informal mathematical claim and $F_t$ the corresponding formal statement. The claim may come from a manuscript, an upstream comment, or documentation, or an explicitly identified Lean-derived description. Project-level semantic alignment review assesses whether $F_t$ expresses $I_t$ using its dependency context $\mathcal{C}$. Agreement with a supplied description establishes fidelity to a manuscript only when that description has been independently checked against the manuscript. A review instance is
\[
    x = (I_t, F_t, \mathcal{C}).
\]

The context $\mathcal{C}$ collects declarations under the extraction policy in Section~\ref{sec:leantrace}. Let $\operatorname{Dep}(F_t)$ contain the declarations referenced by the elaborated statement. For a project declaration $d$, $\operatorname{Dep}(d)$ follows its type and non-proof definition body and includes constructors for inductive types, structures, and classes, so that their constraints are traversed. Proof bodies are excluded. Referenced external-library declarations are retained as boundary nodes, with $\operatorname{Dep}(d)=\varnothing$ for such nodes. The context is then defined by
\[
    \begin{aligned}
        \mathcal{C}^{(0)} &= \operatorname{Dep}(F_t), \\
        \mathcal{C}^{(k+1)} &= \mathcal{C}^{(k)}
            \cup \bigcup_{d \in \mathcal{C}^{(k)}} \operatorname{Dep}(d), \\
        \mathcal{C} &= \bigcup_{k \geq 0} \mathcal{C}^{(k)}.
    \end{aligned}
\]
Thus, $\mathcal{C}$ is the dependency closure of $F_t$ relative to the extraction policy and library boundary.

The assessment assigns each instance $x$ a label
\[
    y(x) \in
    \{\texttt{aligned},\ \texttt{misaligned},\ \texttt{undetermined}\}.
\]
The labels indicate, respectively, that the formal statement expresses the informal claim that a semantic discrepancy has been identified or that the available evidence is insufficient to decide. The judgment is supported by the relevant declarations and a mathematical explanation. Cases labeled \texttt{undetermined} are reserved for subsequent human review and adjudication.

\subsection{LeanTrace}
\label{sec:leantrace}

LeanTrace constructs a statement dependency graph from an elaborated Lean project.
Starting from $F_t$, the elaborated type of the selected theorem declaration, LeanTrace iteratively collects dependencies through declaration types and non-proof definition bodies.
Newly discovered declarations are processed in turn until no unvisited declarations remain, with library declarations retained as boundary nodes without further expansion.
Theorems and other proof-valued declarations contribute their types, but their proof bodies are neither exposed nor traversed.

For inductive types, structures, and classes, LeanTrace also follows their constructors and traverses the constructor types, collecting dependencies introduced by field types and constraints even when no field projection appears explicitly in the target statement.
Operating on elaborated declarations captures both explicit dependencies and those introduced by elaboration, including instances, field projections, and generated auxiliary declarations.

Recursively expanding Mathlib would introduce a large collection of foundational declarations with complex dependencies.
As a practical choice to keep the extracted context manageable, LeanTrace stops at the library boundary, retaining the names and types of referenced declarations from Mathlib and Lean's core libraries without traversing their dependencies or expanding their implementations.
The dependency closure $\mathcal{C}$ defined above is understood relative to these extraction rules and this boundary.

%The extracted dependency declarations and their dependency relations form the context $\mathcal{C}$ used by MechAlign to assess whether $F_t$ expresses the informal claim $I_t$.

% Table~\ref{tab:declaration-kinds} reports the internal declaration kinds in this context, combining selected definition entries and remaining internal declarations. Counts sum occurrences across task closures, so a declaration shared by multiple roots is counted once per task. Kind labels follow the frozen packages: instances and projections are included under their recorded kinds, rather than counted again as separate categories. External-library nodes are excluded because they provide boundary types rather than recursively extracted project content. Definitions and abbreviations account for much of the retained content, but supporting theorem types and generated declarations also contribute substantially; context size therefore cannot be inferred from the number of paired definitions alone.

\subsection{MechAlign}
\label{sec:agent}
MechAlign is an LLM-assisted framework comprising a \emph{Selector}, the LeanTrace extraction tool, and a \emph{Reviewer}.
The Selector uses supplied mappings between informal claims and Lean declarations when available. Otherwise, it searches the manuscript text and Lean source files within a configured scope to identify the manuscript's main results and select corresponding declarations, supporting each selection with quotations from both sources.
For each target, it records the supplied informal claim $I_t$, the selected formal statement $F_t$, and the claim's source.
Claims drawn from Lean comments or documentation are not treated as independently verified manuscript statements.
% The Selector uses explicitly supplied claim-declaration mappings when available. Otherwise, given manuscript text and Lean source files within a configured scope, it identifies the manuscript's main results and selects their corresponding Lean declarations, supported by quotations from both sources. For each target, it obtains the supplied informal claim $I_t$ and the corresponding formal statement $F_t$, recording the claim's source. A claim taken from a Lean comment or documentation is not assumed to be an independently verified manuscript statement.
LeanTrace extracts the dependency context $\mathcal{C}$ for the selected formal statement, retaining declaration types, non-proof values, and dependency relations.

The Reviewer receives context $\mathcal{C}$ together with $I_t$, $F_t$ to assess whether $F_t$ aligns with $I_t$.
It examines six dimensions:
(1) \emph{objects and domains (O\&D)}, checking that the formal types represent the intended mathematical objects and ranges;
(2) \emph{hypotheses (HYP)}, checking for missing, additional, or altered assumptions;
(3) \emph{quantifiers (QUA)}, checking their scope, order, and dependence on previously introduced variables;
(4) \emph{constants and bounds (C\&B)}, checking numerical factors, inequalities, and parameter restrictions;
(5) \emph{conclusions (CON)}, checking that the formal statement covers the claimed result, including all asserted components;
(6) \emph{definition semantics (DEF)}, following retained definitions to check that the underlying constructions and predicates express their intended meanings.

The Reviewer returns an \texttt{aligned}, \texttt{misaligned}, or \texttt{undetermined} judgment, together with explanations for the six dimensions and supporting quotations. Substantive judgments must cite both the supplied informal claim or context and formal evidence. A deterministic evidence check verifies that the quotations occur in the supplied sources; it does not independently validate the informal specification. Ambiguous correspondence or insufficient evidence leads to an \texttt{undetermined} judgment. MechAlign retains selection, extraction, and review records for inspection. The judgments remain model-assisted assessments of the supplied claim rather than formal guarantees of semantic equivalence or manuscript fidelity.

\section{Semantic Alignment of OpenAI Math}
\label{sec:openai-study}

We examine how LeanTrace supports review of a project presented as a proof of a mathematical claim. OpenAI Math provides the primary application.
The study asks whether organizing formal evidence around the selected result helps identify semantic discrepancies and reduces the cost of locating that evidence.
We introduce the review conditions, illustrate dependency inspection through one case, and describe how effectiveness, cost, and limitations are to be compared. Controlled comparative results remain to be completed.

\subsection{Review Setting and Compared Strategies}
\label{sec:review_set}

We used LeanTrace to inspect 201 theorems across 173 projects in \texttt{openai/math}, looking for discrepancies between the stated mathematical claims and their formal encodings.
For each theorem, we compared its informal description in the existing Lean comments with the formal statement and used the extracted dependency graph to inspect the definitions and constraints that determine its meaning.
Upstream docstrings, adjacent comments, and retained paper excerpts provided additional context; generated structural descriptions helped navigate the formal declarations.
This preliminary review assesses correspondence with the supplied descriptions, which have not been independently validated against the manuscripts. Retained paper excerpts provide supplementary context; findings about manuscript fidelity remain candidates pending verification against the original mathematical sources.
The source revision used for the review is listed in Appendix~\ref{app:source-versions}.

To evaluate the benefit of dependency-guided review, we design a comparison of three ways of obtaining evidence for the same claim.
\emph{Statement-only review} exposes the main declaration without dependency content.
\emph{Full-project review} permits ordinary source searches and file readings, leaving the reviewer to locate relevant declarations.
\emph{LeanTrace-assisted review} supplies the extracted dependency graph and retained declarations.
The planned comparison holds the model, claim, root, and source revision fixed, with access rules and budgets disclosed.
Each strategy assesses whether the selected formal result expresses the supplied informal claim and reports any discrepancies with its supporting declarations and mathematical explanations.

\subsection{Case Study: A Missing Dimensional Constraint}
\label{sec:case-hyperbolic}

The main theorem of the HyperbolicObstruction manuscript requires a finite connected \emph{two-dimensional} aspherical simplicial complex with a linear disk-filling bound and a fundamental group admitting no geometric $\CAT(0)$ action.
We use the dimensional requirement to illustrate how project-level alignment review proceeds through the declarations that determine a formal statement's meaning.
The selected Lean result has the following signature (proof body omitted):
\begin{leancode}[\lstset{basicstyle=\small\ttfamily}]
    theorem main : MainStatement.{u}
\end{leancode}
A statement-only reviewer cannot determine from this declaration whether the witness is required to be two-dimensional.

% All 24 retained source declarations; selected dependency paths are audited
% in reports/hyperbolic-case-study/figure-edges.json.
\begin{figure}[tbp]
\centering
\resizebox{0.94\linewidth}{!}{%
%\definecolor{traceaccent}{HTML}{A85249}
%\definecolor{tracefill}{HTML}{FCF5F3}
\colorlet{tracenode}{black!30}
\colorlet{traceedge}{black!30}
\colorlet{tracefraction}{black!70}
\begin{tikzpicture}[
    multiline/.style={minimum height=0.98cm},
    declaration/.style={
        draw=blue!50,
        fill=white,
        rounded corners=3pt,
        line width=1pt,
        minimum height=0.7cm,
        align=center,
        inner sep=3pt,
        font=\small\ttfamily},
    focus/.style={
        declaration,
        draw=red!50,
        line width=1.1pt,
        fill=white},
    dependency/.style={
        ->, >=stealth,
        line width=1pt,
        draw=blue!50},
    folded/.style={
        dependency,
        dashed},
    focusedge/.style={
        dependency,
        draw=red!50,
        line width=1.1pt}
]
    \begin{scope}[shift={(-0.5,0)}, scale=0.90, transform shape]
    \node[focus, text width=2.0cm] (root) at (-0.9,-0) {main};
    \node[focus, text width=3.0cm] (statement) at (-0.9,-1.0539) {MainStatement};
    \node[focus, text width=2.8cm] (complex) at (-0.9,-7.68874) {FiniteComplex};
    \node[focus, text width=2.5cm] (aspherical) at (-0.9,-2.40756) {Aspherical};
    \node[focus, text width=2.9cm] (topology) at (-0.9,-3.84279) {instTopological\\SpaceCarrier};
    \node[focus, text width=1.7cm] (carrier) at (-0.9,-5.23655) {Carrier};
    \node[focus, text width=1.5cm] (vertex) at (-0.9,-6.46265) {Vertex};
    \node[declaration, text width=3.3cm] (filling) at (-5.4,-2.40756) {LinearDiskFilling};
    \node[declaration, text width=1.8cm] (edgepath) at (-7.4,-3.84279) {EdgePath};
    \node[declaration, text width=1.1cm] (edge) at (-7.4,-5.23655) {Edge};
    \node[declaration, text width=1.35cm] (closed) at (-5.4,-3.84279) {Closed};
    \node[declaration, text width=1.1cm] (disk) at (-3.5,-3.84279) {Disk};
    \node[declaration, text width=2.5cm] (homotopy) at (-3.5,-5.23655) {DiskHomotopy};
    \node[declaration, text width=1.9cm] (move) at (-3.5,-6.46265) {DiskMove};
    \node[declaration, multiline, text width=2.8cm] (noaction) at (3.75,-2.40756) {NoGeometric\\CATZeroAction};
    \node[declaration, text width=2.85cm] (action) at (2.4,-3.84279) {GeometricAction};
    \node[declaration, text width=1.5cm] (catzero) at (5.1,-3.84279) {CATZero};
    \node[declaration, text width=2.0cm] (segment) at (8.4,-5.23655) {IsSegment};
    \node[declaration, text width=2.0cm] (geodesic) at (9.5,-3.84279) {IsGeodesic};
    \node[declaration, text width=1.9cm] (catneg) at (7.2,-3.84279) {CATNegOne};
    \node[declaration, multiline, text width=3.3cm] (compatible) at (8.4,-2.40756) {CompatibleLocally\\CATNegOne};
    \node[declaration, text width=3.0cm] (hyperbolic) at (-10,-2.40756) {WordHyperbolic};
    \node[declaration, text width=2.5cm] (distance) at (-10,-3.84279) {WordDistance};
    \node[declaration, text width=2.2cm] (length) at (-10,-5.23655) {WordLength};

    \draw[focusedge] (root.south) -- (statement.north);
    \draw[dependency] (statement.south) -- (filling.north);
    \draw[dependency] (statement.south) -- (noaction.north);
    \draw[dependency] (statement.south) -- (hyperbolic.north);
    \draw[focusedge] (statement.south) -- (aspherical.north);
    \draw[dependency] (statement.south) -- (compatible.north);
    \draw[focusedge] (aspherical.south) -- (topology.north);
    \draw[focusedge] (topology.south) -- (carrier.north);
    \draw[focusedge] (carrier.south) -- (vertex.north);
    \draw[focusedge] (vertex.south) -- (complex.north);
    \draw[dependency] (filling.south) -- (edgepath.north);
    \draw[folded] (edgepath.south) -- (edge.north);
    \draw[dependency] (filling.south) -- (closed.north);
    \draw[dependency] (filling.south) -- (disk.north);
    \draw[dependency] (disk.south) -- (homotopy.north);
    \draw[folded] (homotopy.south) -- (move.north);
    \draw[dependency] (noaction.south) -- (action.north);
    \draw[dependency] (noaction.south) -- (catzero.north);
    \draw[dependency] (geodesic.south) -- (segment.north);
    \draw[dependency] (compatible.south) -- (catneg.north);
    \draw[dependency] (compatible.south) -- (geodesic.north);
    \draw[dependency] (hyperbolic.south) -- (distance.north);
    \draw[dependency] (distance.south) -- (length.north);
    \draw[dependency] (catzero.south) -- (segment.north);
    \draw[dependency] (catneg.south) -- (segment.north);

    \end{scope}

    % The shaded sector is exactly 24/4550 of the circle (1.898901 degrees).
    % Move the entire inset, including its labels, with these coordinates.
    \def\treex{4.6}
    \def\treey{-6.4}
    \begin{scope}[shift={(\treex,\treey)}]
        \draw[draw=black!30, fill=white, rounded corners=3pt, line width=1pt]
            (-1.2,-0.9) rectangle (4.3,0.9);
        \fill[black!10] (0,0) circle (0.72cm);
        \fill[black!70] (0,0) -- (90:0.72cm)
            arc[start angle=90,end angle={90-360*24/4550},radius=0.72cm] -- cycle;
        \fill[white] (0,0) circle (0.51cm);
        \node[font=\small\sffamily\bfseries, text=black!75]
            at (0,0) {0.53\%};
        \node[font=\small\sffamily, align=left, anchor=west,
            text=black!75, text width=3.1cm] at (0.98,0)
            {\textbf{24 / 4,550}\\source declarations};
    \end{scope}
\end{tikzpicture}%
}
\caption{LeanTrace context for HyperbolicObstruction. All 24 retained source
 declarations are shown with selected dependency edges; dashed arrows fold paths
 through auxiliary or constructor nodes. Red nodes and arrows mark the path
 from \texttt{main} to \texttt{FiniteComplex} through the asphericity and carrier
 definitions, inspected to determine whether the witness type enforces the
 manuscript's two-dimensionality requirement. The shaded sector in the inset represents the retained fraction:
 24 of 4,550 source declarations in the main theorem's project import closure
 (0.53\%), counting the root in both.}
\label{fig:hyperbolic-dependencies}
\end{figure}
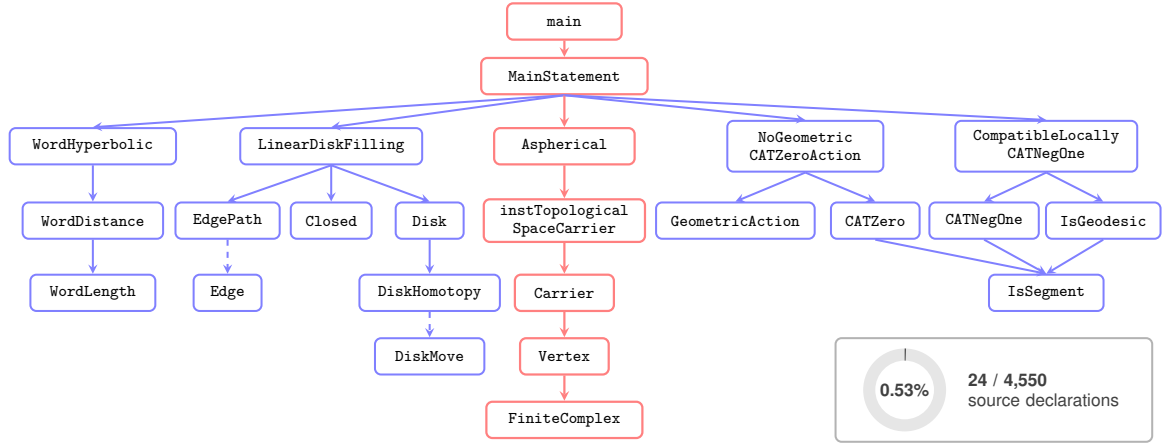

Figure~\ref{fig:hyperbolic-dependencies} shows all 24 retained source declarations and selected dependency paths extracted by LeanTrace.
Following the root to \texttt{MainStatement} exposes the existential witness; the declaration prefix below omits the conjunction of conditions after the quantifiers. The complete definition is reproduced in Appendix~\ref{app:hyperbolic-evidence}:
\begin{leancode}[\lstset{basicstyle=\small\ttfamily}]
    def MainStatement : Prop :=
      ∃ K : FiniteComplex, ∃ x : K.Carrier, ∃ C : ℝ,
        -- Remaining conditions omitted here.
\end{leancode}
The absence of a dimension condition in this excerpt is not sufficient evidence of a missing requirement: that requirement could be encoded in the witness type or in one of the omitted predicates.
Inspecting the retained witness type exposes its complete structure declaration:
\begin{leancode}[\lstset{basicstyle=\small\ttfamily}]
    structure FiniteComplex where
      n : ℕ
      complex : AbstractSimplicialComplex (Fin n)
\end{leancode}
Here \texttt{n} specifies the size of the finite vertex index type; it does not specify the dimension of the complex.
Neither field restricts simplex size to encode two-dimensionality.

The reviewer must also inspect the conditions imposed by \texttt{MainStatement}.
The retained definition of \texttt{Carrier} gives the barycentric realization; \texttt{Aspherical} requires vanishing higher homotopy groups, and \texttt{LinearDiskFilling} bounds the triangle count in a disk certificate in terms of the length of a fillable closed edge path.
Following its dependencies reaches the disk moves, including the triangular move:
\begin{leancode}[\lstset{basicstyle=\small\ttfamily}]
    | triangle (l r) (a b c) (h : {a, b, c} ∈ K.complex.faces)
        (hne : ({a, b, c} : Finset K.Vertex).card = 3) :
        DiskMove K (l ++ [a, b, c] ++ r) (l ++ [a, c] ++ r) 1
\end{leancode}
This constructor restricts an individual disk move to a triangular face; it does not prohibit higher-dimensional simplices in \texttt{K}.
The remaining clauses concern the fundamental group and compatible metrics on homotopy-equivalent finite models.
Inspecting these definitions localizes the discrepancy: the selected proposition does not explicitly require the witness to be two-dimensional.
The complete retained mathematical definitions are supplied in Appendix~\ref{app:hyperbolic-evidence}, so the assessment does not rely on the abbreviated snippets alone.

Although OpenAI's official scope notes explicitly state that the witness's two-dimensionality is not formalized~\citep{openai2026hyperbolic_scope}, an exploratory review using Astra in Codex with full-project access incorrectly reported alignment with the informal claim, focusing on proof details and overlooking this omission. Especially, for this example, LeanTrace retains 24 of the 4,550 source declarations ($0.53\%$) in the upstream main theorem's project import closure, which spans 417 files. These counts include the root theorem and anonymous instances but exclude constructor branches, fields, generated declarations, and external libraries. When constructors and generated declarations are included, the dependency graph contains 41 internal dependencies beyond the root and 152 external-library boundary nodes.

Figure~\ref{fig:hyperbolic-dependencies} displays the retained source declarations with selected edges and folded paths. The red path leads to the witness type, while the disk-filling branch supports the triangle-move check. LeanTrace thus provides a compact context for the Reviewer's assessment of alignment, narrowing the set of declarations to inspect.

\section{LeanReviewBench: A Reusable Review Resource}
\label{sec:benchmark}

The OpenAI Math study illustrates how reviewing a main result requires interpreting the definitions and constraints behind its formal statement.
To make such reviews repeatable across systems, we construct \emph{\bench{}}, a reusable collection of the review instances defined in Section~\ref{sec:task}.
Each instance is a fixed triple $x=(I_t,F_t,\mathcal{C})$: a supplied informal claim, its selected formal statement, and the dependency context extracted by LeanTrace.
The Selector and LeanTrace stages of MechAlign are completed during benchmark construction.
Evaluation begins with the resulting triple and assesses the Reviewer's semantic judgment, with the target correspondence and extracted context held fixed across systems.

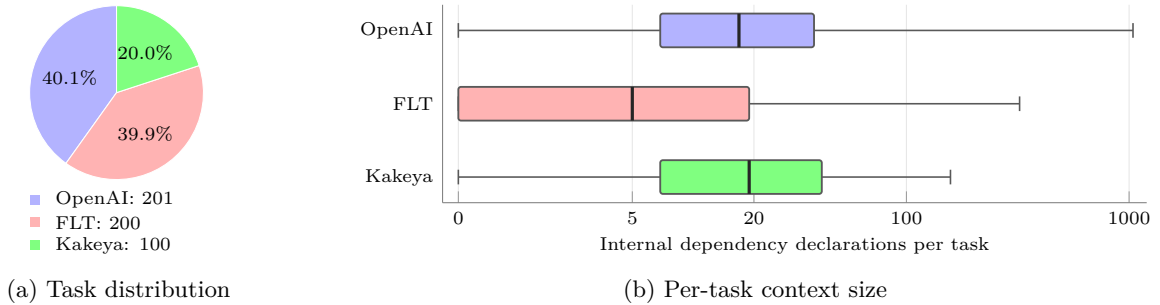
\begin{figure}[t]
\centering
\begin{subfigure}{0.32\linewidth}
\centering
\begin{tikzpicture}[font=\scriptsize]
\pgfmathsetmacro{\openaiend}{90+201/501*360}
\pgfmathsetmacro{\fltend}{90+401/501*360}
\path[fill=blue!30,draw=white] (0,0) -- (90:1.15) arc (90:\openaiend:1.15) -- cycle;
\path[fill=red!30,draw=white] (0,0) -- (\openaiend:1.15) arc (\openaiend:\fltend:1.15) -- cycle;
\path[fill=green!50,draw=white] (0,0) -- (\fltend:1.15) arc (\fltend:450:1.15) -- cycle;
\node at (-0.62,0.20) {40.1\%};
\node at (0.38,-0.54) {39.9\%};
\node at (0.38,0.53) {20.0\%};
\fill[blue!30] (-1.15,-1.48) rectangle +(0.13,0.13);
\node[anchor=west] at (-0.93,-1.415) {OpenAI: 201};
\fill[red!30] (-1.15,-1.78) rectangle +(0.13,0.13);
\node[anchor=west] at (-0.93,-1.715) {FLT: 200};
\fill[green!50] (-1.15,-2.08) rectangle +(0.13,0.13);
\node[anchor=west] at (-0.93,-2.015) {Kakeya: 100};
\end{tikzpicture}
\subcaption{Task distribution}
\end{subfigure}
\hfill
\begin{subfigure}{0.66\linewidth}
\centering
\begin{tikzpicture}
\begin{axis}[
 width=\linewidth,height=4.2cm,
 xmode=log,log basis x=10,
 xmin=0.85,xmax=1200,ymin=0.65,ymax=3.35,
 xtick={1,6,21,101,1001},xticklabels={0,5,20,100,1000},
 ytick={1,2,3},yticklabels={Kakeya,FLT,OpenAI},
 tick label style={font=\scriptsize},
 label style={font=\scriptsize},
 xlabel={Internal dependency declarations per task},
 xlabel style={yshift=2pt},
 axis lines*=left,
 axis line style={black!50},
 tick style={black!40},
 xmajorgrids=true,grid style={black!10},
 ytick style={draw=none},
 clip=false]
% Plot n+1 on a logarithmic axis so zero-sized contexts remain visible.
% Quartiles use linear interpolation (statistics.quantiles, method=inclusive).
% openai: min=0, Q1=7, median=17, Q3=38, max=1041
\draw[black!65,line width=0.6pt] (axis cs:1,3) -- (axis cs:1042,3);
\draw[black!65,line width=0.6pt] (axis cs:1,2.90) -- (axis cs:1,3.10);
\draw[black!65,line width=0.6pt] (axis cs:1042,2.90) -- (axis cs:1042,3.10);
\path[fill=blue!30,draw=black!65,line width=0.7pt,rounded corners=1pt] (axis cs:8.0,2.77) rectangle (axis cs:39.0,3.23);
\draw[black!85,line width=1.2pt] (axis cs:18.0,2.77) -- (axis cs:18.0,3.23);
% flt: min=0, Q1=0, median=5, Q3=19, max=323
\draw[black!65,line width=0.6pt] (axis cs:1,2) -- (axis cs:324,2);
\draw[black!65,line width=0.6pt] (axis cs:1,1.90) -- (axis cs:1,2.10);
\draw[black!65,line width=0.6pt] (axis cs:324,1.90) -- (axis cs:324,2.10);
\path[fill=red!30,draw=black!65,line width=0.7pt,rounded corners=1pt] (axis cs:1.0,1.77) rectangle (axis cs:20.0,2.23);
\draw[black!85,line width=1.2pt] (axis cs:6.0,1.77) -- (axis cs:6.0,2.23);
% kakeya: min=0, Q1=7, median=19, Q3=41.25, max=158
\draw[black!65,line width=0.6pt] (axis cs:1,1) -- (axis cs:159,1);
\draw[black!65,line width=0.6pt] (axis cs:1,0.90) -- (axis cs:1,1.10);
\draw[black!65,line width=0.6pt] (axis cs:159,0.90) -- (axis cs:159,1.10);
\path[fill=green!50,draw=black!65,line width=0.7pt,rounded corners=1pt] (axis cs:8.0,0.77) rectangle (axis cs:42.25,1.23);
\draw[black!85,line width=1.2pt] (axis cs:20.0,0.77) -- (axis cs:20.0,1.23);
\end{axis}
\end{tikzpicture}
\subcaption{Per-task context size}
\end{subfigure}
\caption{\bench{} source composition and context-size distributions. Left: shares of 501 tasks. Right: per-task internal dependency counts, combining selected definitions and other internal declarations. Boxes span the first to third quartile, thick lines mark medians, and whiskers span the minimum to maximum. The horizontal axis uses a logarithmic transformation of count plus one to display zeros and the long tail; tick labels show original counts. Main statements and external-library boundary nodes are excluded.}
\label{fig:source-distribution}
\end{figure}

As illustrated in Figure~\ref{fig:source-distribution}(a), the resource contains 501 tasks: the 201 OpenAI Math main-result tasks discussed in Section~\ref{sec:openai-study}, 200 targets selected from the Fermat's Last Theorem (FLT) development, and 100 from the three-dimensional Kakeya development.
For FLT and Kakeya, we select individual theorem declarations throughout the developments, including intermediate results and supporting mathematical infrastructure, and use each selected declaration as the root for a separate LeanTrace extraction.
FLT targets span elementary and algebraic number theory, local ramification, Galois representations, elliptic curves, modular forms, and modularity lifting.
Kakeya targets cover geometric configurations, shading and density, covering and scale estimates, Frostman-type conditions, and intermediate lemmas leading to dimension bounds.
This selection yields multiple review instances from each development, broadening the mathematical and dependency-structure coverage beyond the OpenAI main-result setting.
%The supplementary targets are not uniformly paired with independently authored informal claims; their description provenance is recorded as part of construction.

To construct each instance, we pin the upstream source and interpretation environment, select a root declaration, and prepare a statement project that retains definition and instance values and structure constraints while replacing theorem proofs with placeholders.
LeanTrace extracts the root's retained dependency graph to form $\mathcal{C}$.
An explicit declaration mapping associates $F_t$ with its supplied description and qualifications, which together specify $I_t$.
We freeze the triple together with provenance and available annotations; reference labels remain separate from the model input.
FLT descriptions were recovered from Lean-generated documentation, while Kakeya uses upstream comments and explicitly marked model-authored Lean paraphrases, with the source breakdown given in Appendix~\ref{app:benchmark-details}.
Compilation records, graph checks, and release validation check artifact integrity without certifying semantic alignment.
Appendix~\ref{app:benchmark-details} gives construction, mapping, validation, and detailed statistics.

% Input from main.tex; requires tikz and pgfplots.
% Context size = selected definitions + remaining internal declarations.
% Excludes the main statement and external-library boundary nodes.

Figure~\ref{fig:source-distribution}(b) characterizes the resulting formal contexts. OpenAI's internal dependency counts are strongly right-skewed: the median is 17 declarations, and the middle half of tasks spans 7–38, while the largest context contains 1,041 internal declarations. FLT has a median of 5 internal declarations, whereas Kakeya has a median of 19 and an upper quartile of 41.25. FLT's first quartile is zero: 64 tasks have no internal dependencies beyond the root, but all retain external-library boundary nodes. Counts exclude the main statement and library-boundary declarations, so they measure retained project dependencies rather than total context or mathematical difficulty. These different context profiles support analysis of reviewer behavior and cost across both small and large dependency closures.

For each triple, a reviewer assesses whether $F_t$ expresses $I_t$ using $\mathcal{C}$ and returns an \texttt{aligned}, \texttt{misaligned}, or \texttt{undetermined} verdict, supported by localized issues, relevant declarations, and a mathematical explanation. Evaluation should report judgment accuracy and answer coverage alongside execution success, model usage, latency, and evidence acquisition cost, separating execution failures from semantic abstentions. Results should be grouped by collection and description provenance: agreement with Lean-derived descriptions measures supplied-text consistency. Released reference labels are provisional model judgments; accuracy against human-adjudicated references remains to be established.

\section{Discussion and Conclusion}
\label{sec:limitations}

We present MechAlign for reviewing whether a selected Lean result expresses its supplied informal claim through the project declarations that determine its meaning. LeanTrace extracts and organizes these dependencies, enabling the Reviewer to investigate semantic discrepancies and cite supporting evidence. Our preliminary OpenAI Math review and HyperbolicObstruction case illustrate this process, while LeanReviewBench provides 501 fixed review instances from OpenAI Math, Fermat's Last Theorem, and Kakeya for reproducible studies of evidence acquisition and semantic judgment.

\textbf{Limitations and Future Works.} The extracted context excludes proof bodies and stops at the external-library boundary, so it may leave essential interpretation evidence unavailable; alignment review also does not certify the original proofs or their axiomatic assumptions. Some informal descriptions are Lean-derived, reference labels remain provisional, and shared developments and possible training exposure limit generalization claims. Future work will validate descriptions against independent mathematical sources, obtain human-adjudicated judgments, and complete controlled comparisons of statement-only, full-project, and dependency-guided reviews to assess judgment quality, evidence localization, and cost.

\section*{Acknowledgment}
This paper is supported by the Strategic Priority Research Program of CAS Grants XDA0480501 and XDA0480502, NSFC Grants 12288201, National Key R\&D Program of China 2023YFA1009401.

% Keep the existing bibliography interface when the manuscript bundle is available.
\IfFileExists{refs.bib}{%
  \IfFileExists{KLMM/klmm.bst}{\bibliographystyle{KLMM/klmm}}{\bibliographystyle{plain}}
  \bibliography{refs}
}{%
  % TODO: supply refs.bib and verify the citations in the untouched introduction
  % and related work before submission.
}
\clearpage
\appendix

\section{Source Revisions and Interpretation Environment}
\label{app:source-versions}

The upstream sources are pinned to the revisions recorded in the
released project metadata. All frozen task packages use the shared
Lean and Mathlib interpretation environment listed below.

\par\smallskip
\begingroup
\small
\renewcommand{\arraystretch}{1.12}
\noindent
\begin{tabularx}{\linewidth}
    {@{}l>{\raggedright\arraybackslash}X@{}}
\toprule
\textbf{Component} & \textbf{Version or revision} \\
\midrule
\multicolumn{2}{@{}l}{\textit{Upstream sources}} \\
\addlinespace[2pt]
\href{https://github.com/openai/math}{OpenAI Math}
    & \nolinkurl{1581d4cbdd69a0faf34b2089614fe2a814316a6a} \\
\href{https://github.com/anthropics/fermats-last-theorem}
    {Fermat's Last Theorem}
    & \nolinkurl{6e837e75355538c7f80bab5b956861e86c4eacc2} \\
\href{https://github.com/project-numina/kakeya-3d}{Kakeya}
    & \nolinkurl{a26427e3c2fcaf9f98694fbd0d4dd49e41dcf5e3} \\
\midrule
\multicolumn{2}{@{}l}{\textit{Interpretation environment}} \\
\addlinespace[2pt]
Lean
    & \nolinkurl{leanprover/lean4:v4.34.1} \\
Mathlib
    & \nolinkurl{d13f23b723b8a846827a245b89c10fc7d3f11612} \\
\bottomrule
\end{tabularx}
\par
\endgroup
% \smallskip

\section{HyperbolicObstruction: Complete Statement Context}
\label{app:hyperbolic-evidence}

Section~\ref{sec:case-hyperbolic} displays only the code needed to explain the dimensional check.
Case~\ref{case:hyperbolic} reproduces the contextual manuscript claim and the retained mathematical definitions, including the complete \texttt{MainStatement} and the constructor branches for disk homotopies and disk moves.
Imports, namespaces, and notation provide the original context; theorem proof bodies and elaborator-generated auxiliary declarations are omitted.
%The declaration-count audit is recorded in \nolinkurl{reports/hyperbolic-case-study/summary.json} and can be reproduced with \nolinkurl{scripts/hyperbolic_case_stats.py} using the pinned upstream sources.

~

% \par
\begingroup
\captionsetup{type=reviewcase}
\tcbset{breakable}
    \begin{informalbox}
    % {HyperbolicObstruction: Contextual Informal Claim}
There exist a finite connected two-dimensional aspherical simplicial complex $K$ and a constant $C<\infty$ with the following properties.
\begin{enumerate}
\item Every combinatorial edge loop of length $\ell$ in the universal cover of $K$ bounds a disk diagram with at most $C\ell$ triangular faces.
\item The group $\pi_1K$ admits no geometric action on a proper complete $\CAT(0)$ space.
\end{enumerate}
In particular, $\pi_1K$ is word-hyperbolic, and no finite simplicial complex homotopy equivalent to $K$ admits a locally $\CAT(0)$ geodesic length metric inducing its given topology. Consequently, no such model admits a compatible locally $\CAT(-1)$ geodesic length metric.
\end{informalbox}

\begin{formalbox}
% {HyperbolicObstruction: Root and Selected Formal Dependencies}
\begin{leancode}
import Mathlib

namespace OAI

noncomputable section

open scoped Topology ContinuousMap

namespace HyperbolicObstruction

structure FiniteComplex where
  n : ℕ
  complex : AbstractSimplicialComplex (Fin n)

namespace FiniteComplex

abbrev Vertex (K : FiniteComplex) := Fin K.n

def Carrier (K : FiniteComplex) :=
  {x : K.Vertex → ℝ // (∀ i, 0 ≤ x i) ∧ (∑ i, x i) = 1 ∧
    ∃ s ∈ K.complex.faces, ∀ i, i ∉ s → x i = 0}

instance (K : FiniteComplex) : TopologicalSpace K.Carrier :=
  inferInstanceAs (TopologicalSpace (Subtype _))

def Edge (K : FiniteComplex) (a b : K.Vertex) : Prop :=
  {a, b} ∈ K.complex.faces

def EdgePath (K : FiniteComplex) : List K.Vertex → Prop
  | [] => False
  | [_] => True
  | a :: b :: t => K.Edge a b ∧ K.EdgePath (b :: t)

def Closed (K : FiniteComplex) (p : List K.Vertex) : Prop :=
  ∃ a t, p = a :: (t ++ [a]) ∨ p = [a]

inductive DiskMove (K : FiniteComplex) : List K.Vertex → List K.Vertex → ℕ → Prop
  | stutter (l r) (a) :
      DiskMove K (l ++ [a, a] ++ r) (l ++ [a] ++ r) 0
  | spur (l r) (a b) (h : K.Edge a b) :
      DiskMove K (l ++ [a, b, a] ++ r) (l ++ [a] ++ r) 0
  | triangle (l r) (a b c) (h : {a, b, c} ∈ K.complex.faces)
      (hne : ({a, b, c} : Finset K.Vertex).card = 3) :
      DiskMove K (l ++ [a, b, c] ++ r) (l ++ [a, c] ++ r) 1

inductive DiskHomotopy (K : FiniteComplex) : List K.Vertex → List K.Vertex → ℕ → Prop
  | refl (p) : DiskHomotopy K p p 0
  | move {p q n} : DiskMove K p q n → DiskHomotopy K p q n
  | symm {p q n} : DiskHomotopy K p q n → DiskHomotopy K q p n
  | trans {p q r m n} : DiskHomotopy K p q m → DiskHomotopy K q r n →
      DiskHomotopy K p r (m + n)

def Disk (K : FiniteComplex) (p : List K.Vertex) (faces : ℕ) : Prop :=
  ∃ a, K.DiskHomotopy p [a] faces

def LinearDiskFilling (K : FiniteComplex) (C : ℝ) : Prop :=
  ∀ p, K.EdgePath p → K.Closed p → (∃ a, K.Disk p a) →
    ∃ a, K.Disk p a ∧ (a : ℝ) ≤ C * (p.length - 1 : ℕ)

def Aspherical (K : FiniteComplex) (x : K.Carrier) : Prop :=
  ∀ n : ℕ, Subsingleton (HomotopyGroup.Pi (n + 2) K.Carrier x)

end FiniteComplex

def IsSegment {X : Type*} [MetricSpace X] (γ : ℝ → X) (x y : X) : Prop :=
  γ 0 = x ∧ γ 1 = y ∧
  ∀ s ∈ Set.Icc (0 : ℝ) 1, ∀ t ∈ Set.Icc (0 : ℝ) 1,
    dist (γ s) (γ t) = |s - t| * dist x y

def IsGeodesic (X : Type*) [MetricSpace X] : Prop :=
  ∀ x y : X, ∃ γ, IsSegment γ x y

def CATZero (X : Type*) [MetricSpace X] : Prop :=
  ∀ x y : X, ∃ γ, IsSegment γ x y ∧
    ∀ z : X, ∀ t ∈ Set.Icc (0 : ℝ) 1,
      dist z (γ t) ^ 2 ≤ (1 - t) * dist z x ^ 2 + t * dist z y ^ 2 -
        t * (1 - t) * dist x y ^ 2

def CATNegOne (X : Type*) [MetricSpace X] : Prop :=
  ∀ x y : X, ∃ γ, IsSegment γ x y ∧
    ∀ z : X, Real.cosh (dist z (γ (1 / 2))) ≤
      (Real.cosh (dist z x) + Real.cosh (dist z y)) /
        (2 * Real.cosh (dist x y / 2))

structure GeometricAction (G X : Type*) [Group G] [MetricSpace X] where
  act : G →* (X ≃ᵢ X)
  proper : ∀ C : Set X, IsCompact C →
    {g : G | ∃ x ∈ C, act g x ∈ C}.Finite
  cocompact : ∃ C : Set X, IsCompact C ∧ ∀ x : X, ∃ g : G, ∃ y ∈ C, act g y = x

universe u

def NoGeometricCATZeroAction (G : Type*) [Group G] : Prop :=
  ∀ (X : Type u) (_ : MetricSpace X), Nonempty X → ProperSpace X → CompleteSpace X →
    CATZero X → IsEmpty (GeometricAction G X)

def WordLength {G : Type*} [Group G] (S : Finset G) (g : G) : ℕ :=
  sInf {n | ∃ w : List G, w.length = n ∧
    (∀ a ∈ w, a ∈ S ∨ a⁻¹ ∈ S) ∧ w.prod = g}

def WordDistance {G : Type*} [Group G] (S : Finset G) (g h : G) : ℝ :=
  WordLength S (g⁻¹ * h)

def WordHyperbolic (G : Type*) [Group G] : Prop :=
  ∃ S : Finset G, Subgroup.closure (S : Set G) = ⊤ ∧
    ∃ δ : ℝ, 0 ≤ δ ∧ ∀ a b c d : G,
      WordDistance S a c + WordDistance S b d ≤
        max (WordDistance S a b + WordDistance S c d)
          (WordDistance S a d + WordDistance S b c) + 2 * δ

def CompatibleLocallyCATNegOne (K : FiniteComplex) : Prop :=
  ∃ m : MetricSpace K.Carrier,
    m.toUniformSpace.toTopologicalSpace = (inferInstance : TopologicalSpace K.Carrier) ∧
    (letI := m
     IsGeodesic K.Carrier ∧ ∀ x : K.Carrier, ∃ r : ℝ, 0 < r ∧
       CATNegOne (Metric.ball x r))

def MainStatement : Prop :=
  ∃ K : FiniteComplex, ∃ x : K.Carrier, ∃ C : ℝ,
    ConnectedSpace K.Carrier ∧ K.Aspherical x ∧ 0 ≤ C ∧ K.LinearDiskFilling C ∧
    NoGeometricCATZeroAction.{u} (FundamentalGroup K.Carrier x) ∧
    WordHyperbolic (FundamentalGroup K.Carrier x) ∧
    ∀ L : FiniteComplex, Nonempty (K.Carrier ≃ₕ L.Carrier) →
      ¬ CompatibleLocallyCATNegOne L

end HyperbolicObstruction

end OAI
\end{leancode}
\end{formalbox}
% \par
\caption{Informal claim and corresponding  formalization in the HyperbolicObstruction project. The informal statement requires a
two-dimensional witness, whereas the selected formal statement
contains no explicit dimension constraint.}
\label{case:hyperbolic}
\endgroup
% \end{reviewcase}

\section{\bench{}: Construction and Detailed Statistics}
\label{app:benchmark-details}

This appendix details how the review triples in Section~\ref{sec:benchmark} are constructed and released. Target selection and LeanTrace extraction precede evaluation; reviewers consume the frozen claim, formal statement, and dependency context without repeating these stages.

\subsection{Target Selection and Coverage}
\label{sec:extension}

The inventory contains 201 OpenAI Math main-result targets, 200 targets from \texttt{anthropics/fermats-last-theorem}, and 100 from \texttt{project-numina/kakeya-3d}. The OpenAI targets are those used in Section~\ref{sec:openai-study}. For FLT and Kakeya, we select individual theorem declarations across the developments, including intermediate results and supporting infrastructure. Each selected declaration becomes the root of a separate review triple. Other theorem declarations reached during extraction contribute their types as context; their occurrence in a dependency closure does not create additional tasks.

The selection covers eight areas in FLT and seven in Kakeya (Table~\ref{tab:directions}). FLT targets range from elementary and algebraic number theory through local ramification, Galois representations, elliptic curves, modular forms, modular-curve geometry, and modularity lifting. Kakeya targets cover geometric configurations, shading and density, covers and scales, Frostman, Katz--Tao, and Sticky estimates, and intermediate lemmas and bootstrap arguments leading to dimension bounds. The counts document the selected coverage, rather than a representative sample of Lean mathematics. OpenAI's domain labels use a different granularity and should not be pooled with these areas into a single domain count.

% \begin{table}[t]
% \centering
% \caption{Composition of the released review-task inventory. Each task has one selected formal statement.}
% \label{tab:sources}
% \begin{tabular}{lrr}
% \toprule
% Source & Tasks & Task share \\
% \midrule
% OpenAI Math & 201 & 40.1\% \\
% Fermat's Last Theorem & 200 & 39.9\% \\
% Kakeya & 100 & 20.0\% \\
% \midrule
% Total & 501 & 100.0\% \\
% \bottomrule
% \end{tabular}
% \end{table}

\begin{table}[t]
\centering
\small
\caption{Selected review targets by area in FLT and Kakeya.}
\label{tab:directions}
\begin{subtable}[t]{0.49\linewidth}
\centering
\subcaption{Fermat's Last Theorem}
\label{tab:directions-flt}
\begin{tabular*}{\linewidth}{@{\hspace{10pt}\extracolsep{\fill}} l r @{\hspace{10pt}}}
\toprule
Area & Targets \\
\midrule
Elementary number theory & 15 \\
Algebraic number theory & 30 \\
Local ramification & 25 \\
Galois representations & 25 \\
Frey and elliptic curves & 30 \\
Modular forms and Hecke operators & 25 \\
Modular-curve geometry & 25 \\
Deformation and modularity lifting & 25 \\
\midrule
Total & 200 \\
\bottomrule
\end{tabular*}
\end{subtable}
\hfill
\begin{subtable}[t]{0.49\linewidth}
\centering
\subcaption{Kakeya}
\label{tab:directions-kakeya}
\begin{tabular*}{\linewidth}{@{\hspace{10pt}\extracolsep{\fill}} l r @{\hspace{10pt}}}
\toprule
Area & Targets \\
\midrule
Geometry & 15 \\
Shading and density & 15 \\
Covers and scales & 15 \\
Frostman, Katz--Tao, and Sticky estimates & 20 \\
Main Lemma 1 & 15 \\
Main Lemma 2 & 15 \\
Bootstrap and dimension conclusions & 5 \\
\midrule
Total & 100 \\
\bottomrule
\end{tabular*}
\end{subtable}
\end{table}

\begin{table}[t]
\centering
\footnotesize
\caption{Internal dependency declaration occurrences by recorded kind, excluding main statements and external-library boundary nodes.}
\label{tab:declaration-kinds}
\setlength{\tabcolsep}{3pt}
\begin{tabular*}{\linewidth}{@{\hspace{10pt}\extracolsep{\fill}} l r r r r r r r r r r @{\hspace{10pt}}}
\toprule
Collection & Def. & Structure & Abbrev. & Class & Inductive & Constructor & Recursor & Theorem & Opaque & Total \\
\midrule
OpenAI & 3,382 & 260 & 1,594 & 12 & 67 & 572 & 49 & 2,830 & 1 & 8,767 \\
FLT & 1,039 & 149 & 385 & 39 & 0 & 188 & 0 & 1,604 & 0 & 3,404 \\
Kakeya & 900 & 291 & 730 & 1 & 0 & 292 & 0 & 700 & 0 & 2,914 \\
\bottomrule
\end{tabular*}
\end{table}

\subsection{Review Instance Construction}
\label{sec:construction}
\label{sec:informal}

We pin upstream source revisions and the interpretation environment (Appendix~\ref{app:source-versions}), register the selected declarations and any required local wrappers, and prepare statement artifacts. Target and auxiliary theorem proofs are replaced with placeholders, while definition and instance values, structure constraints, namespaces, and required notation are retained. LeanTrace extracts each selected statement $F_t$ and its dependency context $\mathcal{C}$ under the policy in Section~\ref{sec:method}, including the external-library boundary.

An explicit declaration mapping associates the selected statement with a description and any scope qualifications, forming the supplied informal claim $I_t$. All 501 tasks have nonempty main descriptions. The 200 FLT descriptions were recovered from upstream HTML documentation generated from Lean. Kakeya's 100 preferred descriptions comprise 82 upstream comments (82\%) and 18 explicitly marked model-authored Lean paraphrases (18\%). The latter include the description of \texttt{KakeyaDimensionThree}; the other 17 expand terse comments and retain explicit hypotheses. All six target statements with an explicit StickyFrostman hypothesis preserve it in their preferred descriptions. Original comments and description origins remain in the accompanying informal records.

For 32 Kakeya targets, the records also contain checked, page-linked background from the Guth--Wang--Zahl paper, identified as \texttt{paper\_context}. Such passages provide background rather than replacing the mapped target description or asserting its equivalence with Lean. For example, Main Lemma 1 in the paper has no exponent loss, whereas the selected Lean result requires $\beta<\gamma$. The records distinguish descriptions, qualifications, and paper context through explicit mapping relations; missing auxiliary descriptions remain absent. Agreement with a Lean-derived description supports a supplied-text consistency test, while source-grounded alignment requires a reviewed independent specification.

The resulting triples are frozen in \texttt{tasks.jsonl}, accompanied by Lean text artifacts and provenance-rich informal records. Evaluation inputs are exported from these frozen packages without reparsing the Lean files or rebuilding the extraction environment. Formal declarations and dependency edges determine the retained context. Description origins, review status, preparation metadata, and reference labels are kept outside the exported model input.

\subsection{Context Statistics and Release Validation}
\label{sec:statistics}
\label{sec:validation}

Figure~\ref{fig:source-distribution}(b) reports the distribution of per-task internal dependency counts, combining the selected definition/structure inventory and remaining internal declarations. Counts exclude main statements and external-library boundary nodes; boxes show the first and third quartiles and median, and whiskers show the full observed range.

In FLT, 64 of the 200 tasks (32\%) have no internal dependency declarations beyond the root. Each nevertheless retains between 4 and 130 external-library boundary nodes. Such roots are expressed through library declarations whose identities and types are retained without recursive expansion. A zero internal count therefore does not imply an empty context; a boundary type may still be insufficient if a judgment requires the library implementation.

The frozen packages separately enumerate a selected definition/structure inventory in their \texttt{definitions} field. Summing these entries across tasks gives 3,477 for OpenAI, 1,155 for FLT, and 1,191 for Kakeya, for a total of 5,823 occurrences. These entries are only part of the retained context, not a count of all internal declarations or of unique definitions. Shared declarations can occur in multiple tasks, and all three collections include tasks with no entries in this field. Table~\ref{tab:declaration-kinds} additionally reports retained internal declarations by kind, including definitions, structures, abbreviations, constructors, classes, and supporting theorem types. These statistics characterize context size and composition rather than established semantic difficulty.

Compilation checks statement types and retained definition bodies in the pinned environment; theorem proofs remain placeholders. Archived closure replay checks the extraction record, while the release validator checks file hashes, target inventories, description mappings, graph consistency, and reference/input bindings. These checks establish artifact integrity without certifying proof correctness or semantic alignment.

The release includes provisional model judgments for all 501 tasks; descriptions and semantic references remain pending human adjudication. Repeated evaluations should record task and model identities, execution configuration, budgets, and execution outcomes, retaining collection and provenance breakdowns. Any held-out split must also account for shared source developments and overlapping dependencies.

\subsection{Code and Data Availability}
\label{app:code-availability}

The MechAlign implementation, including the LeanTrace extraction tool, and the LeanReviewBench data and evaluation tools will be made publicly available upon publication at \url{https://github.com/MechMath/MechAlign} and \url{https://github.com/MechMath/PaperLeanBench}, respectively.
The benchmark repository provides the frozen review instances, provenance records, provisional reference labels, and evaluation tools described above.

The released system supports explicitly supplied claim--declaration mappings, fresh dependency extraction, and six-dimension review with quotation checks.
The original-paper locator is limited to pinned OpenAI TeX sources and archived target mappings; general manuscript search and automatic target selection are not implemented by the release.
Benchmark evaluation uses the frozen supplied claim and dependency context without repeating selection or extraction; the MechAlign adapter assesses the main claim and encodes \texttt{undetermined} as \texttt{insufficient\_information} in protocol submissions.

\end{document}